# Gene-Machine, a new search heuristic algorithm

**Alfredo Garcia Woods**
**2013**

**ABSTRACT**

This paper introduces Gene-Machine, an efficient and new search heuristic algorithm, based in the building-block hypothesis [1] [2]. It is inspired by natural evolution, but does not use some of the concepts present in genetic algorithms like population, mutation and generation. This heuristic exhibits good performance in comparison with genetic algorithms, and can be used to generate useful solutions to optimization and search problems.

*Corresponding Author:*

Alfredo Garcia Woods,
Email: alfredogarciaw@gmail.com

## 1. INTRODUCTION

In this paper we define a new general-purpose heuristic algorithm which can be used to solve different optimization problems. The new heuristic is derived from the study of the Building block hypothesis [1] [2].

Building block hypothesis (BBH): A genetic algorithm seeks optimal performance through the juxtaposition of short, low-order, high-performance schemata, called the building blocks.

This new proposed algorithm does not try to imitate the process of evolution, but presents a mechanism that takes some key concepts from it.

It solves some of the limitations found in Genetic Algorithms, like the loss of BuildingBlocks in the population, necessary for the optimal solution.

We will highlight some similarities and differences from Genetic Algorithms.

## 2. SIMILARITIES AND DIFFERENCES

Genetic Algorithms start with a randomly generated population of solutions. From the current population of solutions the better solutions are selected by the selection operator. The selected solutions are processed by applying recombination and mutation operators. [3]

The solution is usually represented with a fixed-length string over a finite alphabet.

GeneMachine uses the chromosome notion, genes and evolution, but it differs from genetic algorithms, in that it does not use mutation, nor population of individuals, neither the notion of generation.

## 3. ALGORITHM

GeneMachine exploits the concept of BuildingBlock [1] [2], and defines it as: a gene in a given position.
The fitness value of a BuildingBlock is the smaller found in all the chromosomes it participated.

In the algorithm we use "inverse fitness", which means that a smaller value is a better solution.
Each BuildingBlock is ordered in a list (Fitness-List), according to their fitness value.
The BuildingBlocks with smaller fitness value will be in the top of the list.
The GeneMachine algorithm is divided in two phases, "Seeding Phase" and "Growing Phase".

3.1 Seeding Phase

All possible BuildingBlocks will be generated to populate the Fitness-List.
For example: Solving the Salesman problem by using GeneMachine, with the variant that the first and last city do not need to be connected (therefore the graph is open).
The following list of genes: A, B, C, D showed in Figure 1.

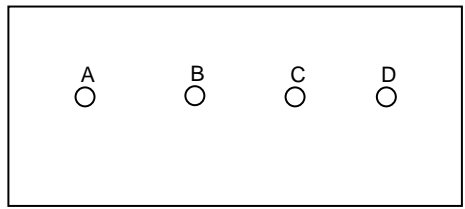

Figure 1 Salesman problem

| Segment | Distance | Segment | Distance | Segment | Distance |
|---|---|---|---|---|---|
| AB | 1 | BC | 1 | CD | 1 |
| AC | 2 | BD | 2 | | |
| AD | 3 | | | | |

Figure 2 Distance table

| Segment | Distance | Segment | Distance | Segment | Distance |
|---|---|---|---|---|---|
| BA | 1 | CB | 1 | DC | 1 |
| CA | 2 | DB | 2 | | |
| DA | 3 | | | | |

Figure 3 Distance table

All possible BuildingBlock are generated.
$A_1, A_2, A_3, A_4, B_1, B_2, B_3, B_4, C_1, C_2, C_3, C_4, D_1, D_2, D_3, D_4$
Note: $A_1$ means gene A in position 1.
For n genes we needed to create $n^2$ BuildingBlocks and n initial Chromosomes.
Note: The chromosomes are not stored, only the BuildingBlocks.
The chromosomes are generated randomly, because in the seeding phase, the Fitness-List is empty.
Example:

Chromosome 1 = $A_1\ C_2\ D_3\ B_4$ fitness value = 5
Chromosome 2 = $B_1\ A_2\ C_3\ D_4$ fitness value = 4
Chromosome 3 = $D_1\ B_2\ A_3\ C_4$ fitness value = 5
Chromosome 4 = $C_1\ D_2\ B_3\ A_4$ fitness value = 4

Then the BuildingBlocks are obtained extracting them of a chromosome, they inherit the fitness value from the chromosome.

| Fitness value | BuildingBlocks |
|---|---|
| 4 | $B_1, A_2, C_3, D_4, C_1, D_2, B_3, A_4$ |
| 5 | $A_1, C_2, D_3, B_4, D_1, B_2, A_3, C_4$ |

Figure 4 Fitness-List

In the seeding phase there is no need to create Chromosomes with all the combinations of genes, but assuring that all the possible BuildingBlocks have been created.

Note: one of the problems of the Genetic Algorithms is the loss of some BuildingBlocks that disappear after successive crossovers, and that they are possibly necessary to find the global optimal solution.

In Gene-Machine this cannot happen because we always have all the possible BuildingBlocks.

### 3.2 Growing Phase

The growing phase has for reason, the promotion of the best BuildingBlocks to upper positions in the Fitness-List.

The combination of the different BuildingBlocks to generate new chromosomes is carried out, by selecting them from the Fitness-List.

The BuildingBlock selection from the Fitness-List will be statistically more frequent for the BuildingBlocks of smaller fitness value. So when a BuildingBlock ascends in the Fitness-List, its probability to be selected increases.

This statistical distribution in the selection frequency is very important, and one of the keys of the algorithm.

The growth phase repeats the process until the time limit is reached.

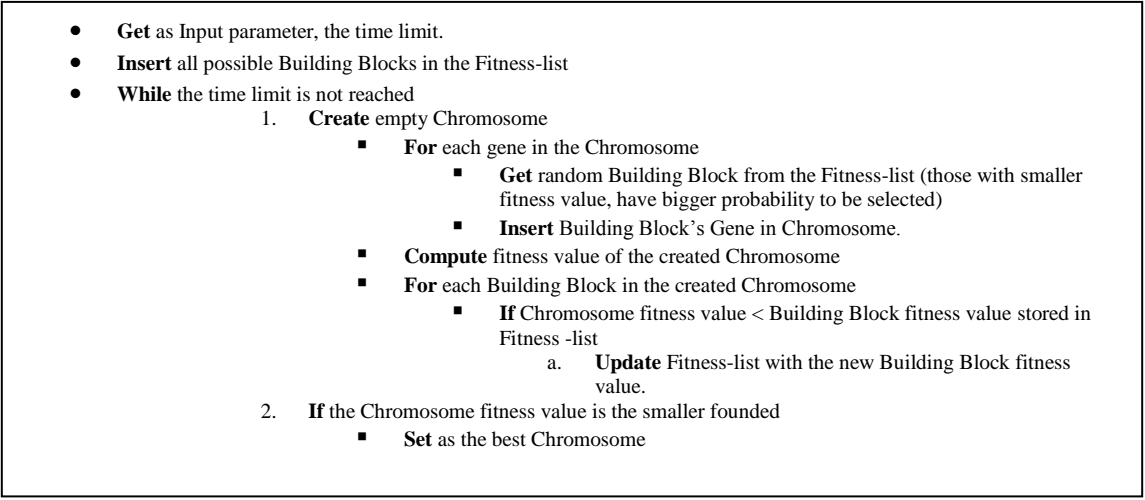

- **Get** as Input parameter, the time limit.
- **Insert** all possible Building Blocks in the Fitness-list
- **While** the time limit is not reached
    1. **Create** empty Chromosome
        - **For** each gene in the Chromosome
            - **Get** random Building Block from the Fitness-list (those with smaller fitness value, have bigger probability to be selected)
            - **Insert** Building Block's Gene in Chromosome.
        - **Compute** fitness value of the created Chromosome
        - **For** each Building Block in the created Chromosome
            - **If** Chromosome fitness value < Building Block fitness value stored in Fitness -list
                a. **Update** Fitness-list with the new Building Block fitness value.
    2. **If** the Chromosome fitness value is the smaller founded
        - **Set** as the best Chromosome

Figure 5 Pseudo code of the Growing phase

The BuildingBlocks that improve (reduce) their fitness value, are promoted to upper positions in the Fitness-List.

| Fitness value | BuildingBlocks |
|---|---|
| 3 | $C_3, A_1, B_2, D_4$ |
| 4 | $B_1, A_2, C_1, D_2, B_3, A_4$ |
| 5 | $C_2, D_3, B_4, D_1, A_3, C_4$ |

Figure 6 Fitness-List

As we mentioned before, the generation notion does not exist, so the time of execution is used to stop the algorithm.

The evolutionary pressure (bigger selection frequency on those BuildingBlocks of smaller fitness value) rises when the time increases.

In the first moments a low evolutionary pressure will be established (not benefitting too much the selection of BuildingBlocks with small fitness value).

The evolutionary pressure is progressively increased when coming closer to the time limit.

The system keeps only one chromosome, the one with the smaller fitness value, to be showed as the best solution found.

### 3.3 Parallel execution

Several Gene-Machines can be executed in parallel, and then merged in a way to avoid local optimal.
To merge two Gene-Machines means to merge their Fitness-Lists.

The process of merging different GeneMachines can be carried out several times.
The demonstration program, available at sourceForge http://sourceforge.net/projects/genemachine/, shows GeneMachine using parallel execution.

```
• Get as Input parameter, the time limit.
• Compute cycle duration = time limit /number of cycles
• Create GeneMachine1 with time limit= cycle duration
• Create GeneMachine2 with time limit= cycle duration
• For cycle=0 to number of cycles
      1. Compute GeneMachine1 evolution
      2. Compute GeneMachine2 evolution
      3. Compute merge Fitness-list from GeneMachine2 in Fitness-list from GeneMachine1
• EndFor
  Print GeneMachine1 best Chromosome
```

Figure 7 Pseudo code of the Parallel execution

```
• Get as Input GeneMachine2
• Get as Input GeneMachine1
• For each BuildingBlock in GeneMachine1. Fitness-list
      1. Get BuildingBlock from GeneMachine2. Fitness-list
      2. If BuildingBlock.fitness value from GeneMachine2. Fitness-list < BuildingBlock. Fitness value from GeneMachine1. Fitness-list
      3. Update GeneMachine1. Fitness-list with BuildingBlock from GeneMachine2. Fitness-list
```

Figure 8 Pseudo code of the Merge mechanism

## 4. CONCLUSION

In this paper, we have explained the Gene-Machine algorithm, which can be used to generate useful solutions to optimization and search problems.

This algorithm has remarkable characteristics, like a very good performance compared with genetic algorithms, and a native capacity of parallelization.

A demonstration of the algorithm is showed at: http://sourceforge.net/projects/genemachine/

The code source is at http://sourceforge.net/p/genemachine/code/2/tree/trunk/